\documentclass[conference]{IEEEtran}
\IEEEoverridecommandlockouts
\usepackage{cite}
\usepackage[symbol]{footmisc}
\usepackage{amsmath,amssymb,amsfonts}
\usepackage{algorithmic}
\usepackage{graphicx}
\usepackage{textcomp}
\usepackage{xcolor}
\usepackage{multirow}
\usepackage{booktabs}

\def\BibTeX{{\rm B\kern-.05em{\sc i\kern-.025em b}\kern-.08em
    T\kern-.1667em\lower.7ex\hbox{E}\kern-.125emX}}
\begin{document}

\title{Hierarchical Scoring for Machine Learning Classifier Error Impact Evaluation
\thanks{Corresponding author: lanus@vt.edu.
}} %

\author{
    \IEEEauthorblockN{
    Erin Lanus \IEEEauthorrefmark{1}, 
    Daniel Wolodkin \IEEEauthorrefmark{2}, %
    Laura J. Freeman \IEEEauthorrefmark{1}\\ %
}
    \IEEEauthorblockA{\IEEEauthorrefmark{1} National Security Institute, Virginia Tech, Arlington, VA, USA} %
    \IEEEauthorblockA{\IEEEauthorrefmark{2} Applied Research Corporation, Virginia Tech, Arlington, VA, USA} %
}

\maketitle

\begin{abstract}
A common use of machine learning (ML) models is predicting the class of a sample.  Object detection is an extension of classification that includes localization of the object via a bounding box within the sample. Classification, and by extension object detection, is typically evaluated by counting a prediction as incorrect if the predicted label does not match the ground truth label. This pass/fail scoring treats all misclassifications as equivalent. In many cases, class labels can be organized into a class taxonomy with a hierarchical structure to either reflect relationships among the data or operator valuation of misclassifications. When such a hierarchical structure exists, hierarchical scoring metrics can return the model performance of a given prediction related to the distance between the prediction and the ground truth label. Such metrics can be viewed as giving partial credit to predictions instead of pass/fail, enabling a finer-grained understanding of the impact of misclassifications. This work develops hierarchical scoring metrics varying in complexity that utilize scoring trees to encode relationships between class labels and produce metrics that reflect distance in the scoring tree. The scoring metrics are demonstrated on an abstract use case with scoring trees that represent three weighting strategies and evaluated by the kind of errors discouraged. Results demonstrate that these metrics capture errors with finer granularity and the scoring trees enable tuning. This work demonstrates an approach to evaluating ML performance that ranks models not only by how many errors are made but by the kind or impact of errors. Python implementations of the scoring metrics will be available in an open-source repository at time of publication.
\end{abstract}

\section{Introduction}
Machine learning (ML) for classification and object detection is increasingly used in artificial intelligence (AI)-enabled systems to support perception functionality. While traditional ML uses algorithms that fit models over engineered features, the power of deep learning to extract features directly from the data via neural network architectures with many layers enables fitting more complex functions with less human processing at the expense of reduced transparency of the model's internals. Test and evaluation (T\&E) approaches that examine the model weights directly are generally not useful, and instead testing typically relies on the use of a test oracle, a set of labeled test samples and a comparison of the model's predictions against the test set. Flat measures for performance evaluation, such as accuracy, precision, recall, and F-measure -- even when computed on a per-class basis -- reflect the pass/fail nature of the oracle. That is, if the ground truth label is ``house cat,'' a prediction of ``house cat'' is considered correct, but any other predicted label is incorrect. When the class labels are nominal and organized into a flat structure, this pass/fail method may adequately capture the expected behavior of the system; however, when the class labels are organized into a hierarchical class structure, all-or-nothing counting over the test set may not provide enough information about system behavior for useful evaluation. Instead, alternative metrics specific to hierarchical classifiers should be used \cite{costa2007review}.

Hierarchical class structures are possible whenever the ground truth labels in the problem domain have unequal relationships whether labels exist at multiple levels in the class structure or not. That is, ``jaguar'' and ``house cat'' are closer in the biological taxonomy than are ``jaguar'' and ``dog'' regardless of the explicit presence in the class set of parent labels ``feline'' and ``canine.'' For this reason, many data sets may be viewed as having hierarchical labels including those representing disparate problem domains such as biological classification, object detection for autonomous vehicles, and medical disease diagnosis. A hierarchical structure to the labels also often implies that errors should not be equally weighted, even if the classifier is not trained with awareness of the structure, and a different hierarchy would imply a different misclassification impact. That is, if the classifier is utilized for a DNA-related problem, calling a ``house cat'' a ``dog'' may result in a worse outcome than calling the same sample a ``jaguar,'' but if the classifier is utilized to detect wildlife on a trail camera for animal control, the opposite would likely be true. Thus, even when the ground truth data does not have an obvious hierarchical structure, the impact of misclassifications may be represented hierarchically \cite{wood2021developing}. 

Hierarchical classification exists in various fields of research. Compared to normal flat classifiers, such as binary or multi-class problems, there exist pre-defined class hierarchies in real-world structures. These class taxonomies are similar to multi-class problems, but exist in a hierarchy with ``IS-A'' relationships\cite{silla2011survey}.

Classification hierarchies are represented by Directed Acyclic Graphs (DAGs) that include a root node with directed paths pointing down to leaf nodes. Most hierarchical classifiers add the requirement that these hierarchies be trees, DAGs where each node only has one parent. Another difference is which classes in the hierarchy are used as labels.  Mandatory leaf-node prediction (MLNP) requires only leaves to be used as labels by the classifier while non-mandatory leaf-node prediction (NMLNP) does not\cite{silla2011survey}.

Various metrics have been considered for evaluating hierarchical classifiers. Historically, flat metrics, such as F-measure, are often used as metrics for classification models. These flat metrics do not properly account for the increased difficulty that comes with predicting deeper levels of a hierarchy. Instead, alternative metrics specific to hierarchical classifiers should be used. These metrics tend to be a combination of four different types of hierarchical metrics\cite{costa2007review}: 

\begin{itemize}
    \item distance-based penalize according to the distance between the true and predicted node but do not address the difficulty of predicting at deeper levels;
    \item depth-dependent penalize errors at higher levels more than those at deeper levels;
    \item semantics-based use features of the individual classes to score classifications, though this is often redundant due to these features being used to build the hierarchy; and
    \item hierarchy-based score according to the ancestral and/or descendant classes of the true and predicted node. 
\end{itemize}

This work develops new hierarchical scoring metrics that utilize scoring trees to addresses drawbacks in current methods. The hierarchical scoring metrics utilize hierarchy-based and distance-based penalties with parameters encoded into scoring trees that allow testers to control depth-dependent penalties as well as introducing the possibility for semantics-based features. The rest of the paper is organized as follows. \S~\ref{background} provides additional background on earlier hierarchical scoring metrics and some of their drawbacks. \S~\ref{metrics} explains the design of new hierarchical scoring metrics. \S~\ref{experiment} describes the experimental setup of an abstract use case for evaluating our proposed metrics against each other with results in \S~\ref{results}. 
\S~\ref{conclusions} concludes the paper.

\section{Background} \label{background}
Early hierarchical scoring measures were hierarchy-based, and created to be similar to the flat F-measure\cite{kiritchenko2005functional}, \cite{kiritchenko2006learning}. For each node $i$, define the set of ancestors of the predicted classification, $\mathcal{P}_i$, as $\hat{\mathcal{P}}_i$ and the set of ancestors of the true classification, $\mathcal{T}_i$, as $\hat{\mathcal{T}}_i$. The intersection of these sets is thought to be similar to the true positive portion of the prediction. This idea is used to find a hierarchical precision (hP) and hierarchical recall (hR) such that:

\begin{equation}
    hP = \dfrac{\sum_i| \hat{\mathcal{P}}_i \cap \hat{\mathcal{T}}_i|}{\sum_i|\hat{\mathcal{P}}_i|} \label{early hP}
\end{equation}

\begin{equation}
    hR = \dfrac{\sum_i| \hat{\mathcal{P}}_i \cap \hat{\mathcal{T}}_i|}{\sum_i|\hat{\mathcal{T}}_i|} \label{early hR}
\end{equation}

These hierarchical metrics are then used in the same manner as flat precision and recall to find a hierarchical F-measure, Such that:
\begin{equation}
    hF_\beta = \dfrac{(\beta^2+1) \times hP \times hR}{(\beta^2 \times hP + hR)}, \beta \in [0, \infty) \label{early hF}
\end{equation}
This metric is used for comparison of hierarchical models  \cite{borges2013evaluation}.  It is recommended for comparing hierarchical classifiers as it is easily applied to all basic hierarchical structures (i.e., trees and DAGs) and scenarios (i.e., MLNP or NMLNP)\cite{silla2011survey}. It is even  recommended for use in more complex scenarios, such as classification problems with multiple labels \cite{cerri2015extensive}.

More recent work compared this metric to pair-based metrics that assign scores to each individual truth/prediction pair. This work found that the existing hF measure over-penalized predictions related to deeper nodes, as the scoring is made based on the number of ancestors, and suggest a similar F-measure based on the lowest common ancestor (LCA) to correct for this. They, however, found that more work is needed to update these scoring methods to include some of the benefits that come with pair-based methods \cite{kosmopoulos2015evaluation}.

To address this, we propose a scoring method similar to the LCA F-measure, with the inclusion of adjustable weighted edges to provide some semantics-based value found in pair-based metrics. While weighted metrics have been criticized for giving uneven scoring to similar pairs in unbalanced trees (trees with various sizes of sub-trees)\cite{cerri2015extensive}, we avoid this issue by standardizing the total edge weight from root to leaf node, and standardize scores according to the depth of the true or predicted class.  

Model predictions with format (truth, prediction) fall into five types for object detection tasks with the first three relevant to classification tasks. 
\begin{enumerate}
    \item ($\mathcal{T}$,$\mathcal{T}$): correct classification of detected object (a type of true positive)
    \item ($\mathcal{T}$,$\mathcal{P}$): incorrect classification of detected object (a type of false positive)
    \item ($\mathcal{P}$,$\mathcal{T}$): incorrect classification of detected object (a type of false negative)
    \item ($\emptyset$,$\mathcal{P}$): ghost detection (a type of false positive)
    \item ($\mathcal{T}$,$\emptyset$): missed detection (a type of false negative)
\end{enumerate}
In the first three types, only the label is incorrect. For the last two types, the location of the prediction also has an error. Ghost detections occur when a model predicts an object where none from the label set exists. Missed detections occur when a model fails to predict an object from the label set that is present. 
\section{Design of New Metrics}\label{metrics}

\begin{figure}
    \centering
    \includegraphics[width=0.75\linewidth]{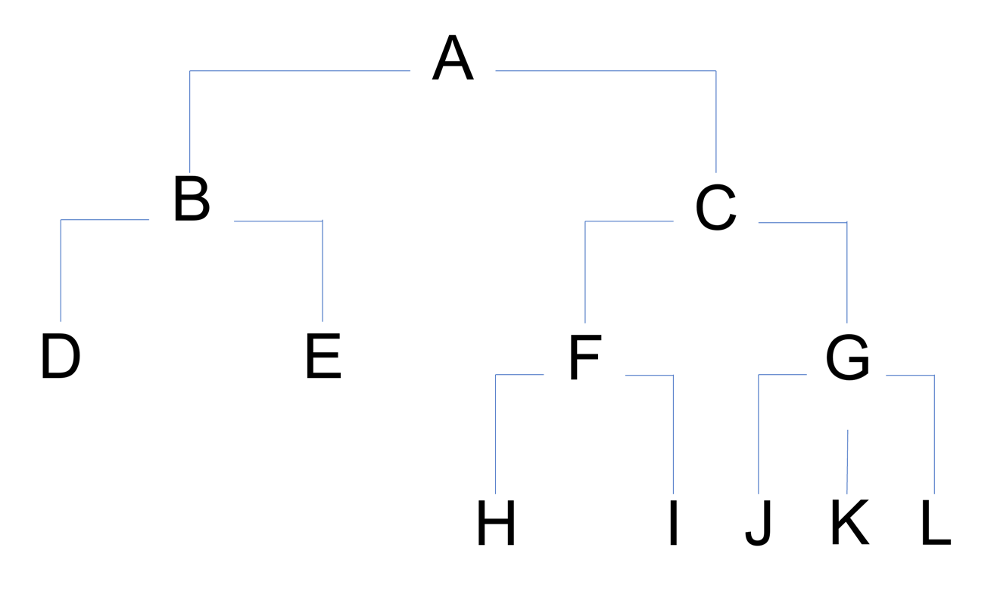}
    \caption{Hierarchical classification trees with root node A.}
    \label{Tree}
\end{figure}

In this section, we develop five metrics for scoring classifiers with labels organized in a tree. An abstract example  used throughout the paper is in Fig.~\ref{Tree}.
The metrics require that the root node, $\mathcal{R}$,  (node A in Fig.~\ref{Tree}) does not appear as a label and is not used as either true $\mathcal{T}$ or predicted $\mathcal{P}$ nodes. 

The first metric is a simple, distance-based measure with a scoring tree with equal weighted edges to use for comparison; equal weighted edges are equivalent to using edge counts. The remaining metrics are hierarchy-based with distance-based penalties with scoring trees with modifiable weighted edges allowing for introduction of depth-dependent scores and providing semantics-based features to penalize different types of errors. The only requirement on the modifiable weights is that all paths from root to leaf have a weight sum of 1.

\subsection{Path Length (PL)}

$\mathbf{PL}$ is a distance-based metric with penalties increasing as predicted labels get further from the true label in the tree. Let $\max(tree)$ be the longest path in the tree and let $d(\mathcal{T},\mathcal{P})$ be the number of edges between $\mathcal{T}$ and $\mathcal{P}$. As the scoring structure must be a tree, the function $d(T,P)$ is unique for all $\mathcal{T},\mathcal{P}$ pairs. Then PL is computed as 
\begin{equation}
    \mathbf{PL}(\mathcal{T},\mathcal{P}) = \dfrac{\max(tree) - d(\mathcal{T},\mathcal{P})}{\max(tree)} \label{PL}.
\end{equation}
When $\mathcal{P}$ is the furthest possible node from  $\mathcal{T}$, $d(\mathcal{T},\mathcal{P}) =\max(tree)$ and $\mathbf{PL}=0$; for example, $\mathbf{PL}(D,L)$ in Fig.~\ref{Tree}.  When $\mathcal{P}=\mathcal{T}$, $d(\mathcal{T},\mathcal{P}) = 0$ and $\mathbf{PL}=1$; for example, $\mathbf{PL}(C,C)$ in Fig.~\ref{Tree}.
 This metric is usable with both MLNP and NMLNP classifiers, giving a perfect score when $\mathcal{P}=\mathcal{T}$ regardless of whether labels are leaf or non-leaf nodes. This metric is simple to use and requires no further information than the tree structure. However, it does not allow for tester control of depth-dependent penalty. For example, it will give the same score when $\mathcal{P}$ is a parent, child, or sibling of $\mathcal{T}$. It has the additional property of being symmetric; that is, $\mathbf{PL}(\mathcal{A},\mathcal{B}) = \mathbf{PL}(\mathcal{B},\mathcal{A})$.

\subsection{Lowest Common Ancestor (L)}
$\mathbf{L}$ computes a depth-based reward summing the weights of edges from  the root to the lowest common ancestor (LCA) of the true and predicted labels. Define the set of edges along the path from $\mathcal{R}$ to $\mathcal{T}$ as $\overrightarrow{\mathcal{T}}$, and the set of edges along the path from $\mathcal{R}$ to $\mathcal{P}$ as $\overrightarrow{\mathcal{P}}$. Then $\mathbf{L}$ is expressed as:
\begin{equation}
    \mathbf{L}(\mathcal{T},\mathcal{P}) = Rew(\mathcal{T},\mathcal{P}) = \sum \{w(e) | e \in (\overrightarrow{\mathcal{T}} \cap \overrightarrow{\mathcal{P}}) \} \label{L}
\end{equation}
This metric is simple, symmetric, and produces scores between 0 and 1 given the constraint that all paths from root to a leaf node in the tree sum to 1. That is, the maximum score is achieved when $\mathcal{T}=\mathcal{P}$ and $\mathcal{T}$ is a leaf node (e.g., $\mathbf{L}(L,L)$ in Fig.~\ref{Tree}), and the minimum score is achieved when the root is the only common ancestor (e.g., $\mathbf{L}(B,C)$ in Fig.~\ref{Tree}),  but insufficient for two reasons. First, it is not suitable for NMLNP; when $\mathcal{T}$ is a non-leaf node, $\mathbf{L}(\mathcal{T},\mathcal{P})$ likely produces a score $<1$ if edge weights are non-zero, even when $\mathcal{T} = \mathcal{P}$, which leads to confusion interpreting the metric where 1 is expected to mean a correct prediction. Additionally, $\mathbf{L}$ does not distinguish between erroneous predictions that occur further down the tree. For example, in Fig.~\ref{Tree}, $\mathbf{L}(H,L) = \mathbf{L}(F,G) = \mathbf{L}(C,C)$.

\subsection{Lowest Common Ancestor with Path Penalty (LPP)}
$\mathbf{LPP}$ adds a distance-based penalty between $\mathcal{T}$ and $\mathcal{P}$ to $\mathbf{L}$'s depth-based reward. The penalty is expressed as:

\begin{equation}
    Pen(\mathcal{T},\mathcal{P}) = -\sum \{w(e) |e \in (\overrightarrow{\mathcal{T}} \cup \overrightarrow{P}) \land e \notin (\overrightarrow{\mathcal{T}} \cap \overrightarrow{\mathcal{P}}) \} \label{Pen}
\end{equation}

When combining reward and penalty, an additional standardization is performed to adjust the metric scale from $[-2,1]$ to  $[0,1]$. The reward portion of the metric produces a score on a $[0,1]$ scale as shown above, but the penalty portion produces scores from $[-2,0]$. That is, there is no penalty when $\mathcal{T}=\mathcal{P}$, but the penalty is -2 when $\mathcal{T}$ and  $\mathcal{P}$ are both leaves and have $\mathcal{R}$ as LCA (e.g., $Pen(\mathcal{D},\mathcal{L})$ in Fig.~\ref{Tree}).

\begin{equation}
    \mathbf{LPP}(\mathcal{T},\mathcal{P}) = \dfrac{Rew(\mathcal{T},\mathcal{P}) + Pen(\mathcal{T},\mathcal{P}) + 2}{3}
\end{equation}

Like $\mathbf{L}$, $\mathbf{LPP}$ is symmetric but not suitable for NMLNP, as correctly predicted leaf nodes will score 1 while correctly predicted nodes higher in the tree will be unable to gain the maximum reward. Two adjustments that standardize scores based on path lengths easily adapt $\mathbf{LPP}$ for suitability with NMLNP problems.

\subsection{Path Standardization}
Lowest Common Ancestor with Path Penalty True Path Standardization $\mathbf{LPP{TPS}}$ standardizes each $\mathbf{LPP{TPS}}(\mathcal{T},\mathcal{P})$ score by the maximum $\mathbf{LPP}$ score possible for the truth node $\mathcal{T}$.
\begin{equation}
    \mathbf{LPP{TPS}}(\mathcal{T},\mathcal{P}) = \dfrac{\mathbf{LPP}(\mathcal{T},\mathcal{P})}{\mathbf{LPP}(\mathcal{T},\mathcal{T})}\label{LPPTPS}
\end{equation}

Lowest Common Ancestor with Path Penalty Predicted Path Standardization $\mathbf{LPP{PPS}}$ standardizes each $\mathbf{LPP{PPS}}(\mathcal{T},\mathcal{P})$ score by the maximum $\mathbf{LPP}$ score possible for the predicted node $\mathcal{P}$.
\begin{equation}
    \mathbf{LPP{PPS}}(\mathcal{T},\mathcal{P}) = \dfrac{\mathbf{LPP}(\mathcal{T},\mathcal{P})}{\mathbf{LPP}(\mathcal{P},\mathcal{P})}\label{LPPTPS}
\end{equation}

Both of these adjusted metrics result in scores of 1 for correct predictions at any level of the hierarchy. However, they are no longer symmetric; $\mathbf{LPP{xPS}}(\mathcal{T},\mathcal{P})$  may not be equal to $\mathbf{LPP{xPS}}(\mathcal{P},\mathcal{T})$. Both metrics could be utilized together similar to equations \ref{early hP} and \ref{early hR} from earlier work in hierarchical scoring \cite{kiritchenko2005functional}, \cite{kiritchenko2006learning}. To do this, define hierarchical Precision ($hP$) and hierarchical Recall ($hR$) per each node $i$ as an average of the standardized scores. First, define $hP_i$ as the average $\mathbf{LPP{PPS}}$ of all $n_i$ classifications where node $i$ is either $\mathcal{T}$ or $\mathcal{P}$. 
\begin{equation}
    hP_i = \dfrac{1}{n_i}\sum \mathbf{LPP{PPS}}(\mathcal{T},\mathcal{P}) \text{ where } \mathcal{T} \text{ or } \mathcal{P} = i  \label{precision}
\end{equation}

Next, define $hR_i$ as the average $\mathbf{LPP{TPS}}$ of all $n_i$ classifications where node $i$ is either $\mathcal{T}$ or $\mathcal{P}$. 
\begin{equation}
    hR_i = \dfrac{1}{n_i}\sum \mathbf{LPP{TPS}}(\mathcal{T},\mathcal{P}) \text{ where } \mathcal{T} \text{ or } \mathcal{P} = i \label{recall}
\end{equation}

Then, similar to equation \ref{early hR},  define the per class F-measure, $hF_{\beta,i}$ using equations \ref{precision} and \ref{recall}. 

\begin{equation}
    hF_{\beta,i} = \dfrac{(\beta^2 + 1)\times hP_i \times hR_i}{\beta^2\times hP_i+hR_i}, \beta \in [0, \infty)
\end{equation}

In this case, consider the unweighted average of each node score $hF_{\beta,i}$ to be the overall model score $hF_\beta$. 

Other compilation methods can be used to assign an overall metric, similar to normal multi-class $F_\beta$ measures. As defined in \cite{sokolova2009systematic} micro-averaging for precision and recall sums the counts of true positives, false positives, and false negatives across classes and then performs division as shown in equations~\ref{microP} and \ref{microR} while macro-averaging computes precision or recall per class and then takes the average as shown in equations~\ref{macroP} and \ref{macroR}. 
\begin{equation}\label{microP}
    P_{\mu} = \dfrac{\sum_{i=1}^{l}tp_i}{\sum_{i=1}^{l}tp_i+fp_i}
\end{equation}
\begin{equation}\label{microR}
    R_{\mu} = \dfrac{\sum_{i=1}^{l}tp_i}{\sum_{i=1}^{l}tp_i+fn_i}
\end{equation}
\begin{equation}\label{macroP}
    P_{M} = \dfrac{\sum_{i=1}^{l} \frac{tp_i}{tp_i+fp_i}}{l}
\end{equation}
\begin{equation}\label{macroR}
    R_{M} = \dfrac{\sum_{i=1}^{l} \frac{tp_i}{tp_i+fn_i}}{l}
\end{equation}

Throughout the remainder of this paper, we consider micro-averaged $\mathbf{LPP{TPS}_{\mu}}$, $\mathbf{LPP{PPS}_{\mu}}$, $hF_{1_{\mu}}$ defined as:
\begin{equation}\label{microhF1}
    hF_{1_{\mu}} = \frac{2* \mathbf{LPP{PPS}_{\mu}} *\mathbf{LPP{TPS}_{\mu}}}{\mathbf{LPP{PPS}_{\mu}}+\mathbf{LPP{TPS}_{\mu}}}.
\end{equation}

\subsection{Modifications for Detection Error}
Thus far, the metrics are designed for evaluating the first three types of errors in \S~\ref{background} for classification problems. In object detection problems, models additionally make the last two types of errors: ghost and missed detections. The metrics described above require modifications to accommodate detection errors. Each modification has strengths and weaknesses. 

One modification is to create a node for $\emptyset$ and place it in the tree such as above $\mathcal{R}$. The strengths of this modification are that the code to implement the scoring metrics extends to treat the $\emptyset$ node like any other in the tree and it is easy to visualize. Additionally, two nodes can be created for $\emptyset_\mathcal{T}$ and $\emptyset_\mathcal{P}$ if differentiating between ghost and missed detections is important. A weakness that makes this modification infeasible is that the scoring metrics are distance-based by design and will result in different scores depending on where the non-$\emptyset$  label resides in tree.

A second modification is to create a node for $\emptyset$ and connect it with edges to every other node. This modification also works with nodes for $\emptyset_\mathcal{T}$ and $\emptyset_\mathcal{P}$. The strengths of this modification are that detection errors are equally weighted regardless of where the non-$\emptyset$ label resides in tree. The weakness that makes this modification not preferred is that the scoring structure is no longer a tree which may lead to unnecessarily complicated computation. Additionally, this adds $n$ (or $2n$) edges to the tree for a label set of size $n$.

The third modification is to score prediction pairs with an $\emptyset$ label consistently. This modification also works with different scores for $\emptyset_\mathcal{T}$ and $\emptyset_\mathcal{P}$. The strengths of this modification are that detection errors are equally weighted regardless of where the non-$\emptyset$ label resides in tree and tree traversal is not needed to compute these scores. The modification requires one extra function in metric implementation to check for and score prediction pairs with an $\emptyset$ label. More importantly, detection errors may reasonably be considered more impactful than any type of classification error, yet the minimum score so far has been based on longest distance in the classification tree. This weakness is countered by adding an offset to raise misclassification error scores to make room at the bottom of the scale for detection errors.

\section{Experimental Setup}\label{experiment}
To evaluate the metrics, we consider predictions made by four abstract models with differing types of misclassification errors, as well as four abstract models  that additionally make detection errors. These also evaluated on a variety of flat and hierarchical metrics with the weighted hierarchical metrics utilizing three example weight strategies on top of the hierarchical label tree in Fig.~\ref{Tree}.

\subsection{Abstract Models}
For experimentation, the abstract test set has 100 samples for each label (non-root) in the tree; that is, it is a collection of 100 copies of each ground truth label. Abstract models are defined by how they behave when predicting on these samples. 
\begin{itemize}
    \item \textbf{Model 1 ``always correct'':} predicts correctly.
    \item \textbf{Model 2 ``very wrong'':} predicts furthest nodes.
    \item \textbf{Model 3 ``cautious'':} predicts a valid node with slight errors that are closer to the root (i.e., a parent or grandparent when available, or else self).
    \item \textbf{Model 4 ``aggressive'':} predicts a valid node with slight errors that are closer to the leaves (e.g., a child or grandchild if available, or else self or a sibling).
\end{itemize}
 Confusion matrices for Models 2-4 in Tables~\ref{model2}-\ref{model4} further demonstrate the model predictions for each  truth label. Model 1 is not shown as its matrix is 100 on the diagonal. 
 Due to the root node not being a valid label, Model 3 is occasionally correct when classifying higher-level nodes. To balance this for comparison, Model 4 is also made to be occasionally correct at the leaf node level so that both models have the same average $F_1$ score.

\begin{table}
    \caption{Model 2 Confusion Matrix}
    \centering
     \begin{tabular}{c|*{11}{c}}
    \textbf{True} &\multicolumn{11}{c}{\textbf{Predicted}}\\
            \hline
        & \textbf{B} & \textbf{C} & \textbf{D} & \textbf{E} & \textbf{F} & \textbf{G} & \textbf{H} & \textbf{I} & \textbf{J} & \textbf{K} & \textbf{L}  \\ \hline 
        \textbf{B} & 0 & 0 & 0 & 0 & 0 & 0 & 0 & 0 & 0 & 0 & 100 \\ 
        \textbf{C} & 0 & 0 & 0 & 100 & 0 & 0 & 0 & 0 & 0 & 0 & 0 \\ 
        \textbf{D} & 0 & 0 & 0 & 0 & 0 & 0 & 0 & 0 & 0 & 100 & 0 \\ 
        \textbf{E} & 0 & 0 & 0 & 0 & 0 & 0 & 0 & 0 & 0 & 100 & 0 \\ 
        \textbf{F} & 0 & 0 & 0 & 100 & 0 & 0 & 0 & 0 & 0 & 0 & 0 \\ 
        \textbf{G} & 0 & 0 & 0 & 100 & 0 & 0 & 0 & 0 & 0 & 0 & 0 \\ 
        \textbf{H} & 0 & 0 & 0 & 100 & 0 & 0 & 0 & 0 & 0 & 0 & 0 \\ 
        \textbf{I} & 0 & 0 & 0 & 100 & 0 & 0 & 0 & 0 & 0 & 0 & 0 \\ 
        \textbf{J} & 0 & 0 & 0 & 100 & 0 & 0 & 0 & 0 & 0 & 0 & 0 \\ 
        \textbf{K} & 0 & 0 & 0 & 100 & 0 & 0 & 0 & 0 & 0 & 0 & 0 \\ 
        \textbf{L} & 0 & 0 & 0 & 100 & 0 & 0 & 0 & 0 & 0 & 0 & 0 \\ 
    \end{tabular}
    \label{model2}
\end{table}

\begin{table}
    \caption{Model 3 Confusion Matrix}
    \centering
     \begin{tabular}{c|*{11}{c}}
    \textbf{True} &\multicolumn{11}{c}{\textbf{Predicted}}\\
            \hline
        & \textbf{B} & \textbf{C} & \textbf{D} & \textbf{E} & \textbf{F} & \textbf{G} & \textbf{H} & \textbf{I} & \textbf{J} & \textbf{K} & \textbf{L}  \\ \hline 
        \textbf{B} & 100 & 0 & 0 & 0 & 0 & 0 & 0 & 0 & 0 & 0 & 0 \\ 
        \textbf{C} & 0 & 100 & 0 & 0 & 0 & 0 & 0 & 0 & 0 & 0 & 0 \\ 
        \textbf{D} & 100 & 0 & 0 & 0 & 0 & 0 & 0 & 0 & 0 & 0 & 0 \\ 
        \textbf{E} & 100 & 0 & 0 & 0 & 0 & 0 & 0 & 0 & 0 & 0 & 0 \\ 
        \textbf{F} & 0 & 100 & 0 & 0 & 0 & 0 & 0 & 0 & 0 & 0 & 0 \\ 
        \textbf{G} & 0 & 100 & 0 & 0 & 0 & 0 & 0 & 0 & 0 & 0 & 0 \\ 
        \textbf{H} & 0 & 50 & 0 & 0 & 50 & 0 & 0 & 0 & 0 & 0 & 0 \\ 
        \textbf{I} & 0 & 50 & 0 & 0 & 50 & 0 & 0 & 0 & 0 & 0 & 0 \\ 
        \textbf{J} & 0 & 50 & 0 & 0 & 0 & 50 & 0 & 0 & 0 & 0 & 0 \\ 
        \textbf{K} & 0 & 50 & 0 & 0 & 0 & 50 & 0 & 0 & 0 & 0 & 0 \\ 
        \textbf{L} & 0 & 50 & 0 & 0 & 0 & 50 & 0 & 0 & 0 & 0 & 0 \\ 
    \end{tabular}
        \label{model3}
\end{table}

\begin{table}[]
    \caption{Model 4 Confusion Matrix}
    \centering
     \begin{tabular}{c|*{11}{c}}
    \textbf{True} &\multicolumn{11}{c}{\textbf{Predicted}}\\
            \hline
        & \textbf{B} & \textbf{C} & \textbf{D} & \textbf{E} & \textbf{F} & \textbf{G} & \textbf{H} & \textbf{I} & \textbf{J} & \textbf{K} & \textbf{L}  \\ \hline 
        \textbf{B} & 0 & 0 & 50 & 50 & 0 & 0 & 0 & 0 & 0 & 0 & 0 \\ 
        \textbf{C} & 0 & 0 & 0 & 0 & 0 & 0 & 20 & 20 & 20 & 20 & 20 \\ 
        \textbf{D} & 0 & 0 & 30 & 70 & 0 & 0 & 0 & 0 & 0 & 0 & 0 \\ 
        \textbf{E} & 0 & 0 & 70 & 30 & 0 & 0 & 0 & 0 & 0 & 0 & 0 \\ 
        \textbf{F} & 0 & 0 & 0 & 0 & 0 & 0 & 50 & 50 & 0 & 0 & 0 \\ 
        \textbf{G} & 0 & 0 & 0 & 0 & 0 & 0 & 0 & 0 & 33 & 33 & 34 \\ 
        \textbf{H} & 0 & 0 & 0 & 0 & 0 & 0 & 28 & 72 & 0 & 0 & 0 \\ 
        \textbf{I} & 0 & 0 & 0 & 0 & 0 & 0 & 72 & 28 & 0 & 0 & 0 \\ 
        \textbf{J} & 0 & 0 & 0 & 0 & 0 & 0 & 0 & 0 & 28 & 36 & 36 \\ 
        \textbf{K} & 0 & 0 & 0 & 0 & 0 & 0 & 0 & 0 & 36 & 28 & 36 \\ 
        \textbf{L} & 0 & 0 & 0 & 0 & 0 & 0 & 0 & 0 & 36 & 36 & 28 \\ 
    \end{tabular}
        \label{model4}
\end{table}

Four models modify Model 3 and Model 4 to represent detection errors with confusion matrices in Tables~\ref{model3b}-\ref{model4c}.
\begin{itemize}
    \item \textbf{Models 3b and 4b:} Models 3 and 4 with 10\% of predictions as missed detections.
    \item \textbf{Models 3c and 4c:} Models 3 and 4 with an extra 10\% predictions as ghost detections with labels selected evenly across all nodes.
\end{itemize}

\begin{table}
    \caption{Model 3b Confusion Matrix}
    \centering
        \begin{tabular}{c|*{12}{c}}
    \textbf{True} &\multicolumn{12}{c}{\textbf{Predicted}}\\ \hline
        & \textbf{B} & \textbf{C} & \textbf{D} & \textbf{E} & \textbf{F} & \textbf{G} & \textbf{H} & \textbf{I} & \textbf{J} & \textbf{K} & \textbf{L}  & \textbf{$\varnothing$} \\ \hline
        \textbf{B} & 90 & 0 & 0 & 0 & 0 & 0 & 0 & 0 & 0 & 0 & 0 & 10 \\ 
        \textbf{C} & 0 & 90 & 0 & 0 & 0 & 0 & 0 & 0 & 0 & 0 & 0 & 10 \\ 
        \textbf{D} & 90 & 0 & 0 & 0 & 0 & 0 & 0 & 0 & 0 & 0 & 0 & 10 \\ 
        \textbf{E} & 90 & 0 & 0 & 0 & 0 & 0 & 0 & 0 & 0 & 0 & 0 & 10 \\ 
        \textbf{F} & 0 & 90 & 0 & 0 & 0 & 0 & 0 & 0 & 0 & 0 & 0 & 10 \\ 
        \textbf{G} & 0 & 90 & 0 & 0 & 0 & 0 & 0 & 0 & 0 & 0 & 0 & 10 \\ 
        \textbf{H} & 0 & 45 & 0 & 0 & 45 & 0 & 0 & 0 & 0 & 0 & 0 & 10 \\ 
        \textbf{I} & 0 & 45 & 0 & 0 & 45 & 0 & 0 & 0 & 0 & 0 & 0 & 10 \\ 
        \textbf{J} & 0 & 45 & 0 & 0 & 0 & 45 & 0 & 0 & 0 & 0 & 0 & 10 \\ 
        \textbf{K} & 0 & 45 & 0 & 0 & 0 & 45 & 0 & 0 & 0 & 0 & 0 & 10 \\ 
        \textbf{L} & 0 & 45 & 0 & 0 & 0 & 45 & 0 & 0 & 0 & 0 & 0 & 10 \\ 
        \textbf{$\varnothing$} & 0 & 0 & 0 & 0 & 0 & 0 & 0 & 0 & 0 & 0 & 0 & 0\\ 
    \end{tabular}
        \label{model3b}
\end{table}

\begin{table}
    \caption{Model 4b Confusion Matrix}
    \centering
        \begin{tabular}{c|*{12}{c}}
    \textbf{True} &\multicolumn{12}{c}{\textbf{Predicted}}\\ \hline
        & \textbf{B} & \textbf{C} & \textbf{D} & \textbf{E} & \textbf{F} & \textbf{G} & \textbf{H} & \textbf{I} & \textbf{J} & \textbf{K} & \textbf{L}  & \textbf{$\varnothing$} \\ \hline
        \textbf{B} & 0 & 0 & 45 & 45 & 0 & 0 & 0 & 0 & 0 & 0 & 0 & 10 \\ 
        \textbf{C} & 0 & 0 & 0 & 0 & 0 & 0 & 18 & 18 & 18 & 18 & 18 & 10 \\ 
        \textbf{D} & 0 & 0 & 27 & 63 & 0 & 0 & 0 & 0 & 0 & 0 & 0 & 10 \\ 
        \textbf{E} & 0 & 0 & 63 & 27 & 0 & 0 & 0 & 0 & 0 & 0 & 0 & 10 \\ 
        \textbf{F} & 0 & 0 & 0 & 0 & 0 & 0 & 45 & 45 & 0 & 0 & 0 & 10 \\ 
        \textbf{G} & 0 & 0 & 0 & 0 & 0 & 0 & 0 & 0 & 30 & 30 & 30 & 10 \\ 
        \textbf{H} & 0 & 0 & 0 & 0 & 0 & 0 & 25 & 65 & 0 & 0 & 0 & 10 \\ 
        \textbf{I} & 0 & 0 & 0 & 0 & 0 & 0 & 65 & 25 & 0 & 0 & 0 & 10 \\ 
        \textbf{J} & 0 & 0 & 0 & 0 & 0 & 0 & 0 & 0 & 25 & 32 & 33 & 10 \\ 
        \textbf{K} & 0 & 0 & 0 & 0 & 0 & 0 & 0 & 0 & 33 & 25 & 32 & 10 \\ 
        \textbf{L} & 0 & 0 & 0 & 0 & 0 & 0 & 0 & 0 & 32 & 33 & 25 & 10 \\ 
        \textbf{$\varnothing$} & 0 & 0 & 0 & 0 & 0 & 0 & 0 & 0 & 0 & 0 & 0 & 0\\ 
    \end{tabular}
        \label{model4b}
\end{table}

\begin{table}
    \caption{Model 3c Confusion Matrix}
    \centering
        \begin{tabular}{c|*{12}{c}}
    \textbf{True} &\multicolumn{12}{c}{\textbf{Predicted}}\\ \hline
        & \textbf{B} & \textbf{C} & \textbf{D} & \textbf{E} & \textbf{F} & \textbf{G} & \textbf{H} & \textbf{I} & \textbf{J} & \textbf{K} & \textbf{L}  & \textbf{$\varnothing$} \\ \hline
        \textbf{B} & 100 & 0 & 0 & 0 & 0 & 0 & 0 & 0 & 0 & 0 & 0 & 0\\ 
        \textbf{C} & 0 & 100 & 0 & 0 & 0 & 0 & 0 & 0 & 0 & 0 & 0 & 0\\ 
        \textbf{D} & 100 & 0 & 0 & 0 & 0 & 0 & 0 & 0 & 0 & 0 & 0 & 0\\ 
        \textbf{E} & 100 & 0 & 0 & 0 & 0 & 0 & 0 & 0 & 0 & 0 & 0 & 0\\ 
        \textbf{F} & 0 & 100 & 0 & 0 & 0 & 0 & 0 & 0 & 0 & 0 & 0 & 0\\ 
        \textbf{G} & 0 & 100 & 0 & 0 & 0 & 0 & 0 & 0 & 0 & 0 & 0 & 0\\ 
        \textbf{H} & 0 & 50 & 0 & 0 & 50 & 0 & 0 & 0 & 0 & 0 & 0 & 0\\ 
        \textbf{I} & 0 & 50 & 0 & 0 & 50 & 0 & 0 & 0 & 0 & 0 & 0 & 0\\ 
        \textbf{J} & 0 & 50 & 0 & 0 & 0 & 50 & 0 & 0 & 0 & 0 & 0 & 0\\ 
        \textbf{K} & 0 & 50 & 0 & 0 & 0 & 50 & 0 & 0 & 0 & 0 & 0 & 0\\ 
        \textbf{L} & 0 & 50 & 0 & 0 & 0 & 50 & 0 & 0 & 0 & 0 & 0 & 0\\ 
        \textbf{$\varnothing$} & 10 & 10 & 10 & 10 & 10 & 10 & 10 & 10 & 10 & 10 & 10 & 0\\
    \end{tabular}
        \label{model3c}
\end{table}

\begin{table}
    \caption{Model 4c Confusion Matrix}
    \centering
        \begin{tabular}{c|*{12}{c}}
    \textbf{True} &\multicolumn{12}{c}{\textbf{Predicted}}\\ \hline
        & \textbf{B} & \textbf{C} & \textbf{D} & \textbf{E} & \textbf{F} & \textbf{G} & \textbf{H} & \textbf{I} & \textbf{J} & \textbf{K} & \textbf{L}  & \textbf{$\varnothing$} \\ \hline
        \textbf{B} & 0 & 0 & 50 & 50 & 0 & 0 & 0 & 0 & 0 & 0 & 0 & 0\\ 
        \textbf{C} & 0 & 0 & 0 & 0 & 0 & 0 & 20 & 20 & 20 & 20 & 20 & 0\\ 
        \textbf{D} & 0 & 0 & 30 & 70 & 0 & 0 & 0 & 0 & 0 & 0 & 0 & 0\\ 
        \textbf{E} & 0 & 0 & 70 & 30 & 0 & 0 & 0 & 0 & 0 & 0 & 0 & 0\\ 
        \textbf{F} & 0 & 0 & 0 & 0 & 0 & 0 & 50 & 50 & 0 & 0 & 0 & 0\\ 
        \textbf{G} & 0 & 0 & 0 & 0 & 0 & 0 & 0 & 0 & 33 & 33 & 34 & 0\\ 
        \textbf{H} & 0 & 0 & 0 & 0 & 0 & 0 & 28 & 72 & 0 & 0 & 0 & 0\\ 
        \textbf{I} & 0 & 0 & 0 & 0 & 0 & 0 & 72 & 28 & 0 & 0 & 0 & 0\\ 
        \textbf{J} & 0 & 0 & 0 & 0 & 0 & 0 & 0 & 0 & 28 & 36 & 36 & 0\\ 
        \textbf{K} & 0 & 0 & 0 & 0 & 0 & 0 & 0 & 0 & 36 & 28 & 36 & 0\\ 
        \textbf{L} & 0 & 0 & 0 & 0 & 0 & 0 & 0 & 0 & 36 & 36 & 28 & 0\\ 
        \textbf{$\varnothing$} & 10 & 10 & 10 & 10 & 10 & 10 & 10 & 10 & 10 & 10 & 10 & 0\\
    \end{tabular}
        \label{model4c}
\end{table}

\subsection{Weight Strategies}
Scoring trees designed to represent three general weighting strategies demonstrate how edge weights effect evaluation. 
\begin{itemize}
        \item \textbf{Decreasing (D):} 90\% of the edge weight placed on edges out of the root, and decreasing toward the leaf nodes.
        \item \textbf{Non-increasing (N):} trade-off equal edge weights both along a path and across a level. 
    \item \textbf{Increasing (I):} 10\% of the edge weight placed on edges out of the root, and increasing toward the leaf nodes.
\end{itemize}

\begin{figure}[t]
\centering
\includegraphics[width=0.75\linewidth]{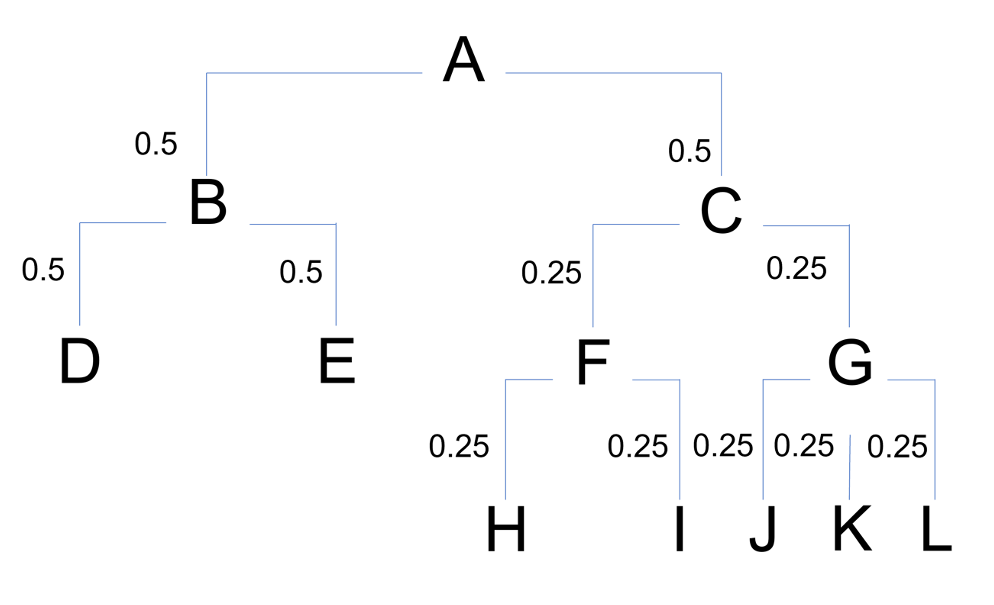}
\caption{Graph of non-increasing edge weight scoring tree}
\label{Basic}
\end{figure}

\begin{figure}[t]
\centering
\includegraphics[width=0.75\linewidth]{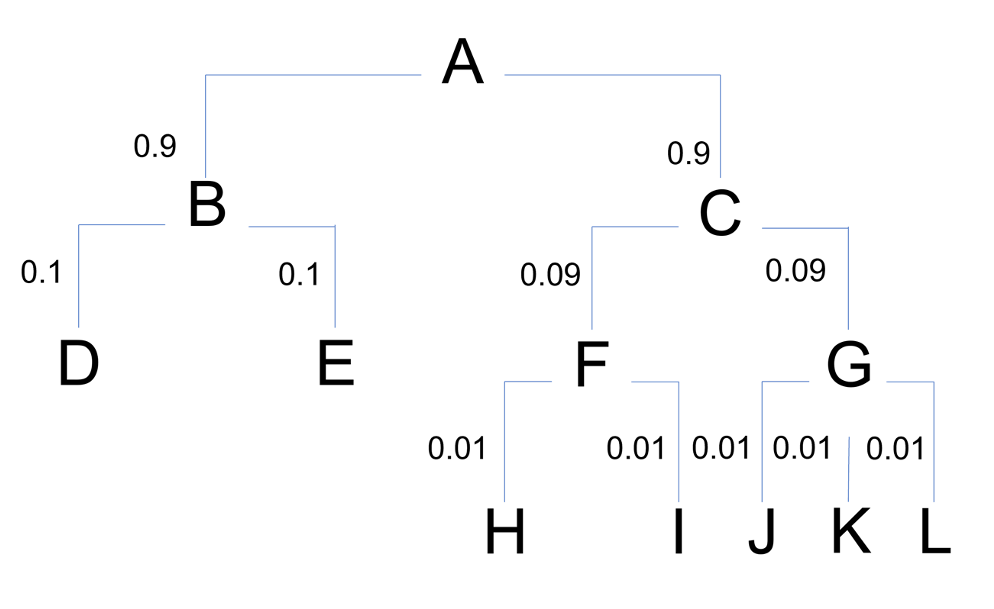}
\caption{Graph of decreasing edge weight scoring tree}
\label{Decreasing}
\end{figure}

\begin{figure}[t]
\centering
\includegraphics[width=0.75\linewidth]{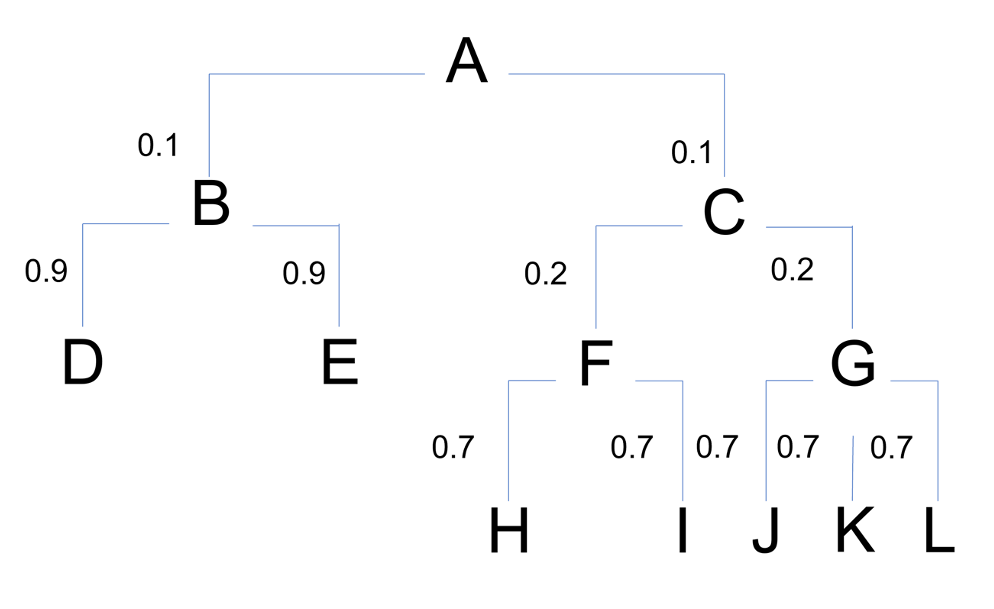}
\caption{Graph of increasing edge weight scoring tree}
\label{Increasing}
\end{figure}

\section{Results}\label{results}
In this section, we evaluate the models on six flat measures, one hierarchical measure that does not use weighting (PL), and five weighted hierarchical measures using the three weight strategies. The flat measures use both macro-averaging ($M$) and micro-averaging ($\mu$) while for hierarchical measures we only report micro-averaged scores. Comparison of macro-averaged hierarchical scores and against previous hierarchical measures is left for future work. 
To see the impact of offset for ghost and missed detection errors, we consider two offsets. Scores where the ghost and missed detection error offset is 0 are in Table~\ref{Scores0} while scores for offset -1  are in Table~\ref{Scores-1}

\begin{table}[]
    \caption{Scores, ghost and missed error = 0}
    \label{Scores0}
    \centering
            \begin{tabular}{l|*{4}{c}*{4}{c}}
            \toprule
    &\multicolumn{8}{c}{\textbf{Model}}\\
        \textbf{Metric} & \textbf{1} & \textbf{2} & \textbf{3} & \textbf{3b} & \textbf{3c} &\textbf{4} & \textbf{4b} & \textbf{4c} \\
        \toprule
$R_M$  & 1.00 & 0.00 & 0.18 & 0.23 & 0.17 & 0.18 & 0.24 & 0.17\\
$P_M$  & 1.00 & 0.73 & 0.68 & 0.63 & 0.13 & 0.48 & 0.44 & 0.18\\
$F1_M$  & 1.00 & 0.00 & 0.07 & 0.07 & 0.07 & 0.14 & 0.13 & 0.13\\
$R_{\mu}$  & 1.00 & 0.00 & 0.18 & 0.16 & 0.17 & 0.18 & 0.17 & 0.17\\
$P_{\mu}$  & 1.00 & 0.00 & 0.18 & 0.16 & 0.17 & 0.18 & 0.17 & 0.17\\
$F1_{\mu}$  & 1.00 & 0.00 & 0.18 & 0.16 & 0.17 & 0.18 & 0.17 & 0.17\\
\midrule
$PL$ & 1.00 & 0.09 & 0.79 & 0.71 & 0.72 & 0.73 & 0.66 & 0.66\\
\midrule
${D-L}$ & 0.98 & 0.00 & 0.92 & 0.83 & 0.84 & 0.96 & 0.88 & 0.88\\
${D-LPP}$ & 0.99 & 0.01 & 0.95 & 0.86 & 0.87 & 0.97 & 0.88 & 0.88\\
${D-{TPS}}$ & 1.00 & 0.03 & 0.96 & 0.87 & 0.87 & 0.98 & 0.89 & 0.89\\
${D-{PPS}}$ & 1.00 & 0.04 & 0.98 & 0.88 & 0.89 & 0.97 & 0.88 & 0.88\\
$D-hF_{\mu}$ & 1.00 & 0.03 & 0.97 & 0.87 & 0.88 & 0.97 & 0.89 & 0.89\\
\midrule
${N-L}$ & 0.86 & 0.00 & 0.56 & 0.50 & 0.51 & 0.72 & 0.65 & 0.65\\
${N-LPP}$ & 0.95 & 0.05 & 0.75 & 0.68 & 0.68 & 0.76 & 0.69 & 0.69\\
${N-{TPS}}$ & 1.00 & 0.19 & 0.84 & 0.76 & 0.76 & 0.83 & 0.75 & 0.75\\
${N-{PPS}}$ & 1.00 & 0.28 & 0.86 & 0.77 & 0.78 & 0.82 & 0.75 & 0.75\\
$N-hF_{\mu}$ & 1.00 & 0.23 & 0.85 & 0.76 & 0.77 & 0.83 & 0.75 & 0.75\\
        \midrule
${I-L}$ & 0.71 & 0.00 & 0.15 & 0.13 & 0.13 & 0.37 & 0.33 & 0.33\\
${I-LPP}$ & 0.90 & 0.10 & 0.53 & 0.47 & 0.48 & 0.46 & 0.42 & 0.42\\
${I-{TPS}}$ & 1.00 & 0.58 & 0.89 & 0.80 & 0.81 & 0.74 & 0.67 & 0.67\\
${I-{PPS}}$ & 1.00 & 0.77 & 0.62 & 0.56 & 0.56 & 0.87 & 0.79 & 0.79\\
$I-hF_{\mu}$ & 1.00 & 0.66 & 0.73 & 0.66 & 0.66 & 0.80 & 0.73 & 0.73\\ 
        \bottomrule
    \end{tabular}
\end{table}

\begin{table}[]
    \caption{Scores, ghost and missed error = -1}
    \label{Scores-1}
    \centering
            \begin{tabular}{l|*{4}{c}*{4}{c}}
            \toprule
    &\multicolumn{8}{c}{\textbf{Model}}\\
        \textbf{Metric} & \textbf{1} & \textbf{2} & \textbf{3} & \textbf{3b} & \textbf{3c} &\textbf{4} & \textbf{4b} & \textbf{4c} \\
        \toprule
$R_M$ & 1.00 & 0.00 & 0.18 & 0.23 & 0.17 & 0.18 & 0.24 & 0.17\\
$P_M$ & 1.00 & 0.73 & 0.68 & 0.63 & 0.13 & 0.48 & 0.44 & 0.18\\
$F1_M$ & 1.00 & 0.00 & 0.07 & 0.07 & 0.07 & 0.14 & 0.13 & 0.13\\
$R_{\mu}$ & 1.00 & 0.00 & 0.18 & 0.16 & 0.17 & 0.18 & 0.17 & 0.17\\
$P_{\mu}$ & 1.00 & 0.00 & 0.18 & 0.16 & 0.17 & 0.18 & 0.17 & 0.17\\
$F1_{\mu}$ & 1.00 & 0.00 & 0.18 & 0.16 & 0.17 & 0.18 & 0.17 & 0.17\\
\midrule
$PL$ & 1.00 & 0.24 & 0.83 & 0.74 & 0.75 & 0.77 & 0.7 & 0.70\\
\midrule
$D-L$ & 0.99 & 0.50 & 0.96 & 0.86 & 0.87 & 0.98 & 0.89 & 0.89\\
$D-LPP$ & 0.99 & 0.25 & 0.97 & 0.87 & 0.88 & 0.98 & 0.89 & 0.89\\
$D-TPS$ & 1.00 & 0.26 & 0.97 & 0.87 & 0.88 & 0.98 & 0.89 & 0.89\\
$D-PPS$ & 1.00 & 0.28 & 0.98 & 0.89 & 0.89 & 0.98 & 0.89 & 0.89\\
$D-hF1_{\mu}$ & 1.00 & 0.27 & 0.98 & 0.88 & 0.89 & 0.98 & 0.89 & 0.89\\
\midrule
$N-L$ & 0.93 & 0.50 & 0.78 & 0.70 & 0.71 & 0.86 & 0.78 & 0.78\\
$N-LPP$ & 0.97 & 0.28 & 0.81 & 0.73 & 0.74 & 0.82 & 0.75 & 0.75\\
$N-TPS$ & 1.00 & 0.35 & 0.87 & 0.78 & 0.79 & 0.86 & 0.79 & 0.79\\
$N-PPS$ & 1.00 & 0.43 & 0.88 & 0.80 & 0.80 & 0.86 & 0.78 & 0.78\\
$N-hF_{\mu}$ & 1.00 & 0.39 & 0.88 & 0.79 & 0.80 & 0.86 & 0.78 & 0.78\\
        \midrule
$I-L$ & 0.85 & 0.5 & 0.57 & 0.52 & 0.52 & 0.68 & 0.62 & 0.62\\
$I-LPP$ & 0.93 & 0.32 & 0.65 & 0.58 & 0.59 & 0.6 & 0.54 & 0.54\\
$I-TPS$ & 1.00 & 0.61 & 0.9 & 0.81 & 0.82 & 0.76 & 0.69 & 0.69\\
$I-PPS$ & 1.00 & 0.79 & 0.65 & 0.58 & 0.59 & 0.88 & 0.80 & 0.80\\
$I-hF1_{\mu}$ & 1.00 & 0.69 & 0.75 & 0.68 & 0.69 & 0.82 & 0.74 & 0.74\\
        \bottomrule
    \end{tabular}
\end{table}

\begin{figure}
    \centering
    \includegraphics[width=0.9\linewidth]{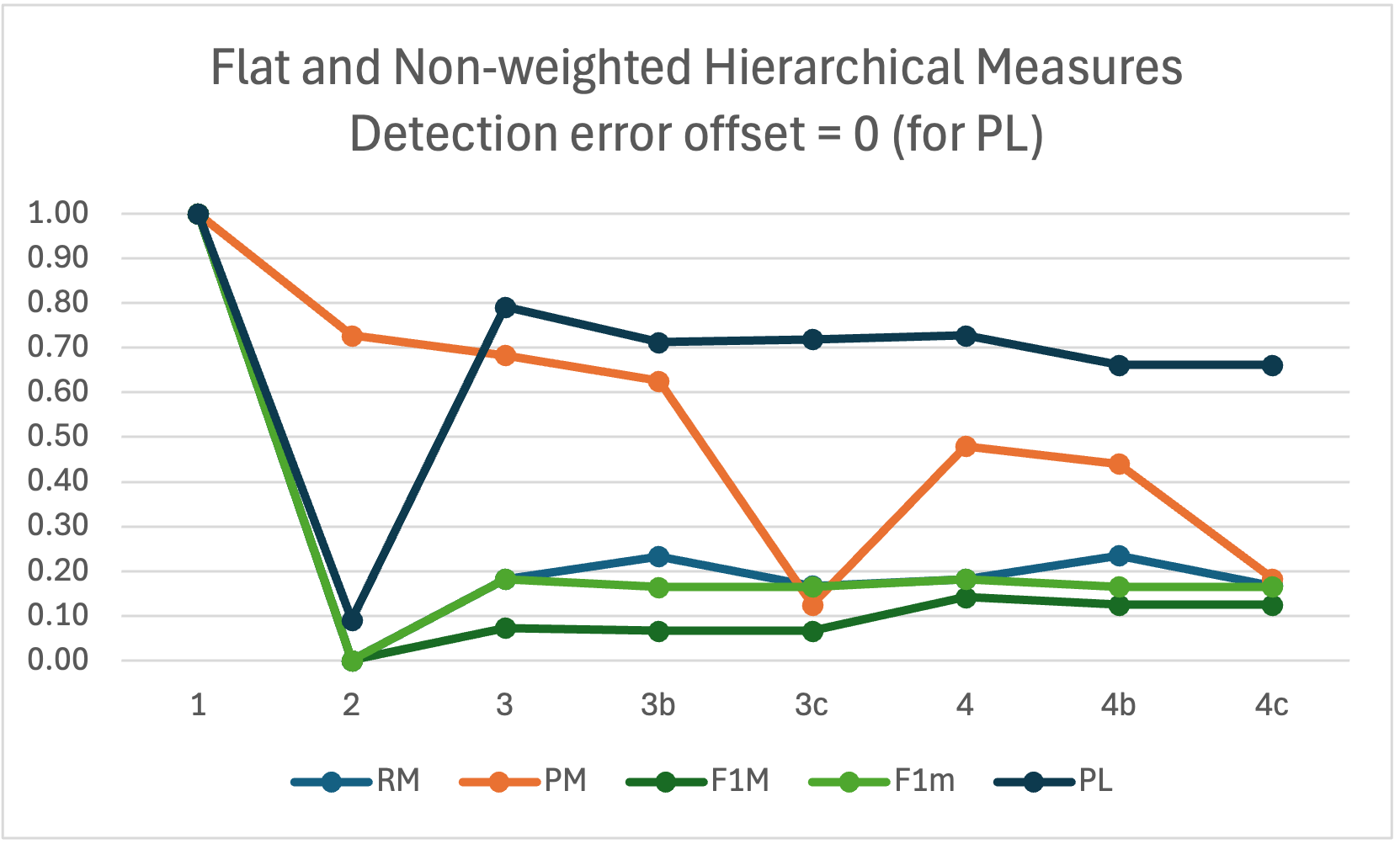}
    \caption{Flat and non-weighted hierarchical measures}
    \label{fig:flatnonweighted}
\end{figure}
\begin{figure}
    \centering
    \includegraphics[width=0.9\linewidth]{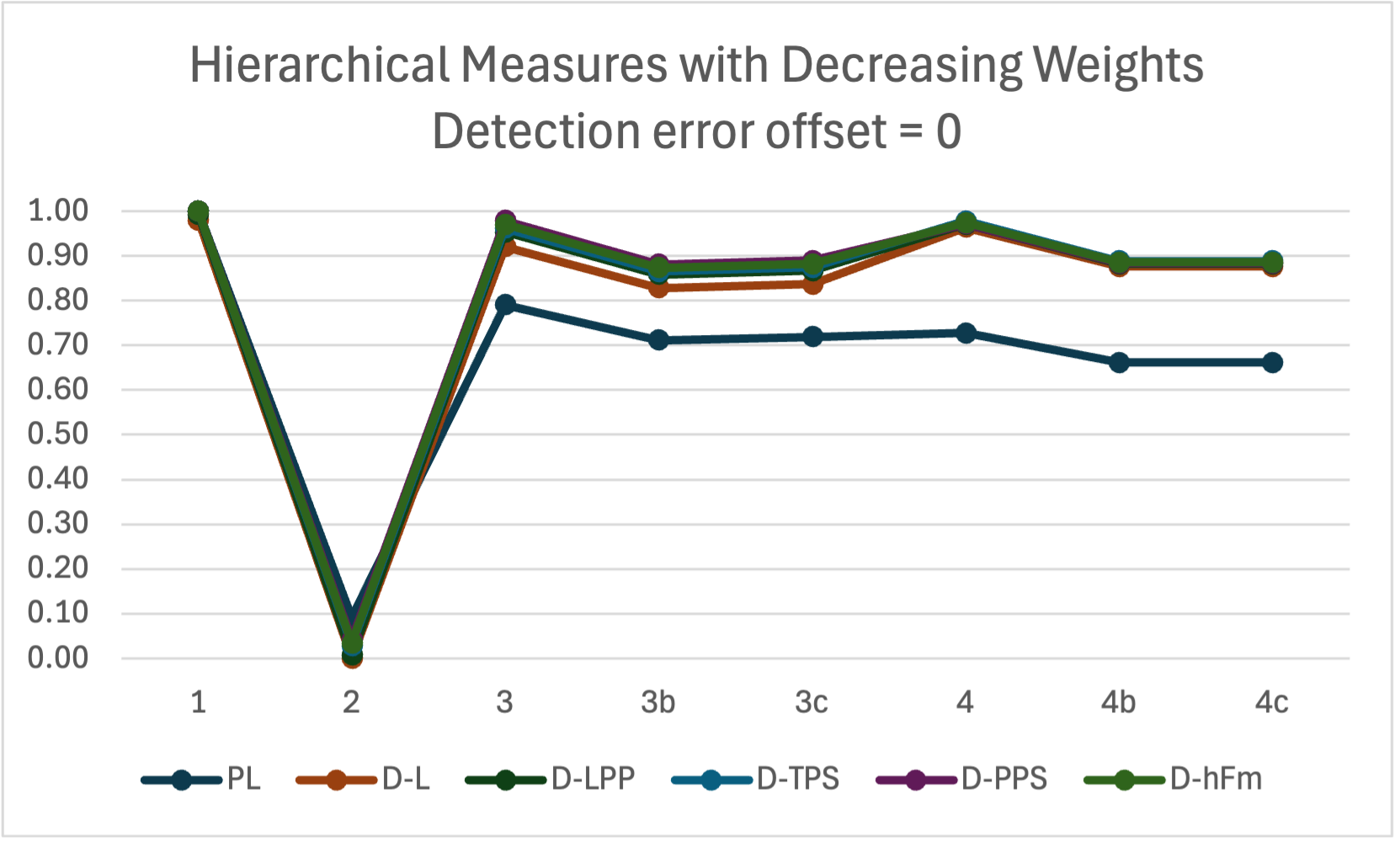}
    \caption{Hierarchical measures with decreasing weights.}
    \label{fig:decreasing}
\end{figure}
\begin{figure}
    \centering
    \includegraphics[width=0.9\linewidth]{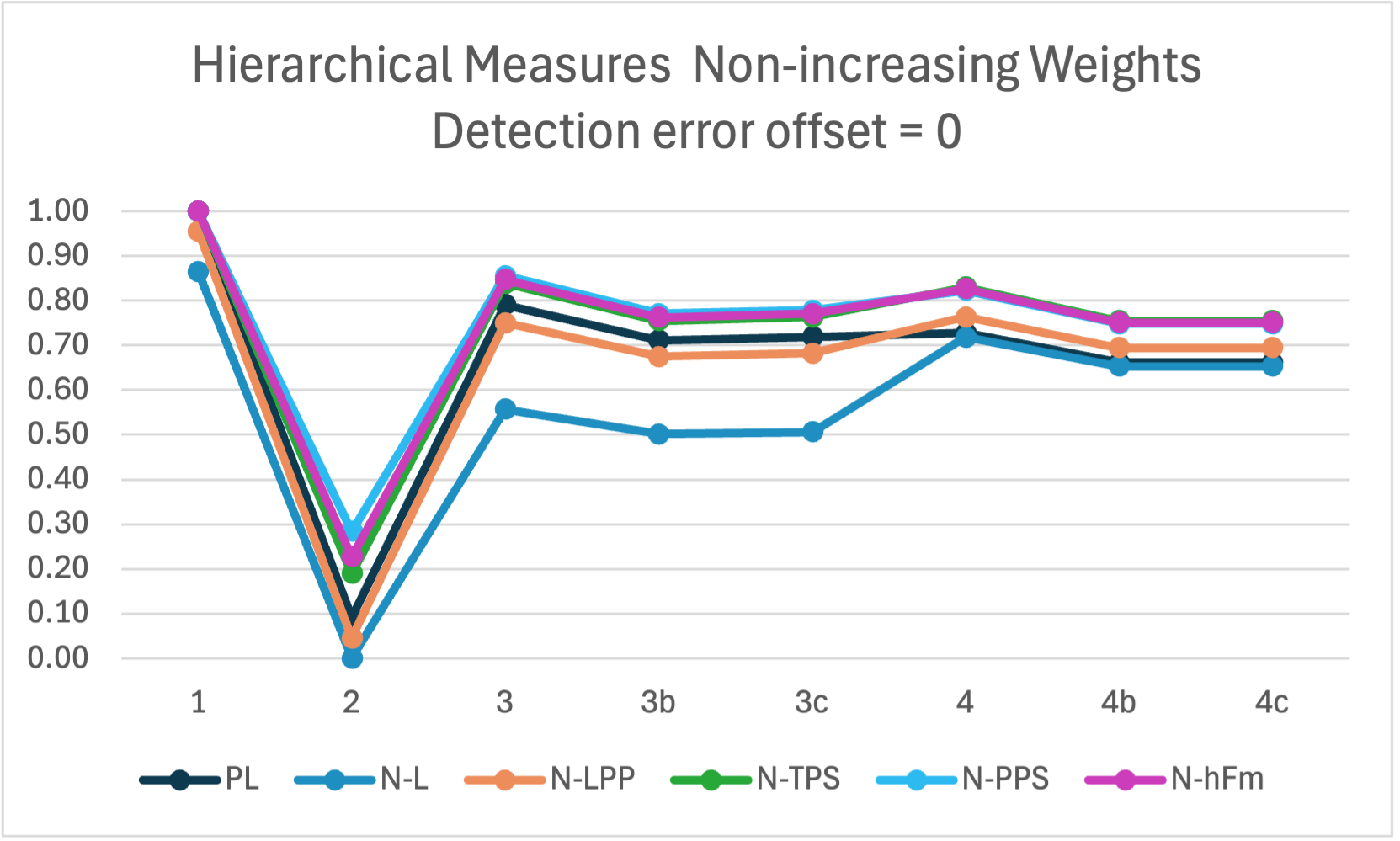}
    \caption{Hierarchical measures with non-increasing weights.}
    \label{fig:nonincreasing}
\end{figure}
\begin{figure}
    \centering
    \includegraphics[width=0.9\linewidth]{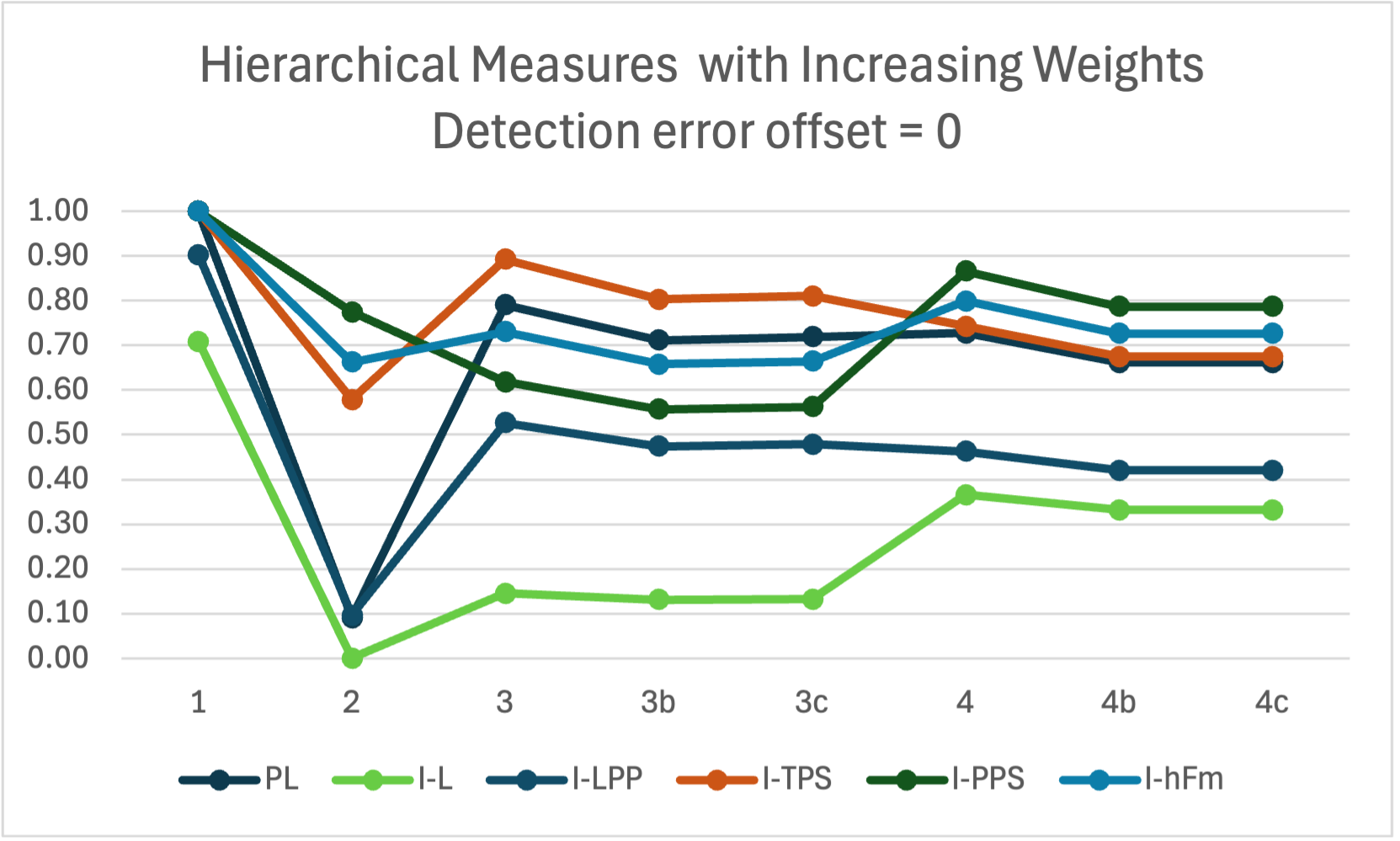}
    \caption{Hierarchical measures with increasing weights.}
    \label{fig:increasing}
\end{figure}

\begin{figure}
    \centering
    \includegraphics[width=0.9\linewidth]{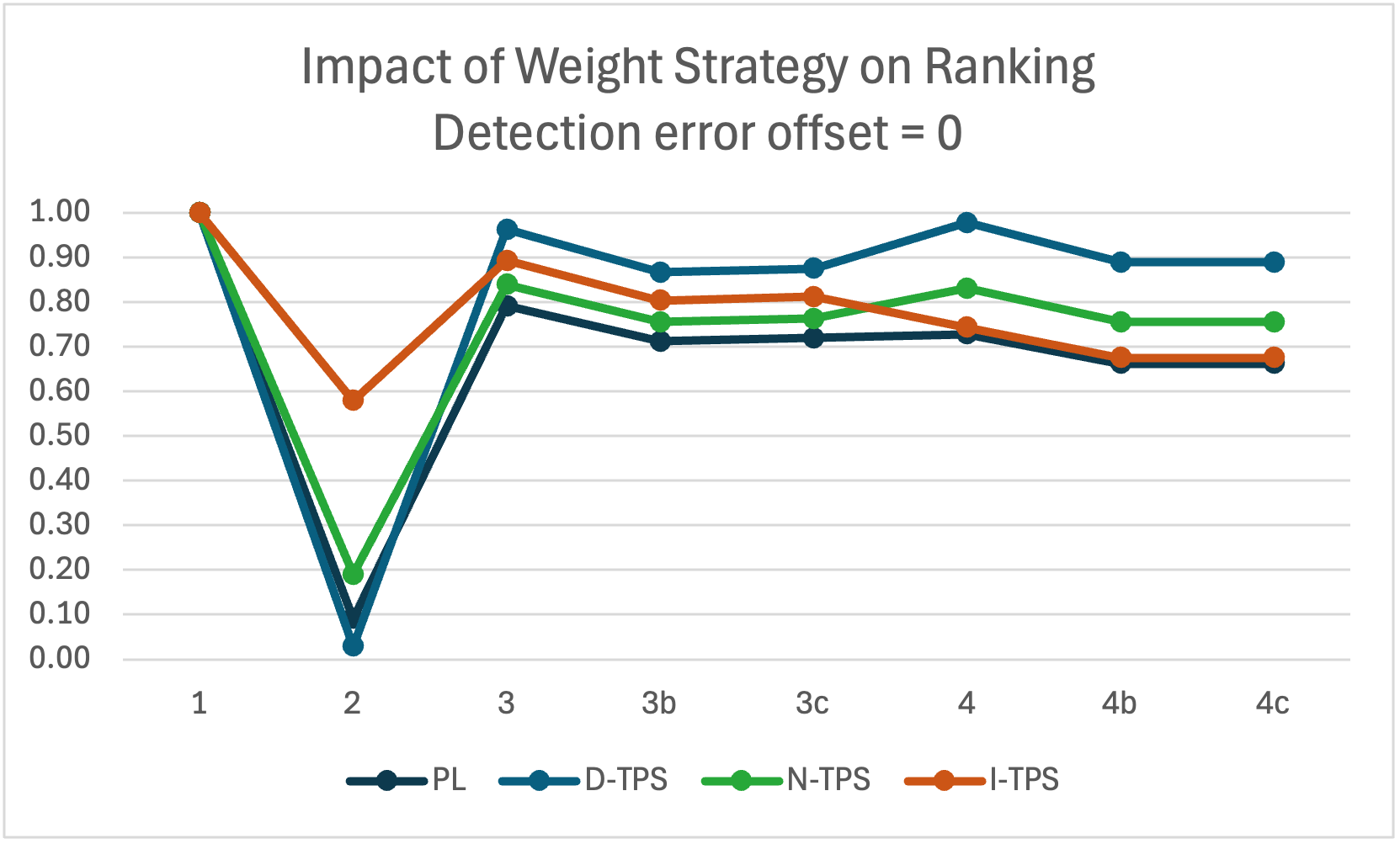}
    \caption{Impact of weight strategy on ranking, detection error offset = 0.}
    \label{fig:impact}
\end{figure}
\begin{figure}
    \centering
    \includegraphics[width=0.9\linewidth]{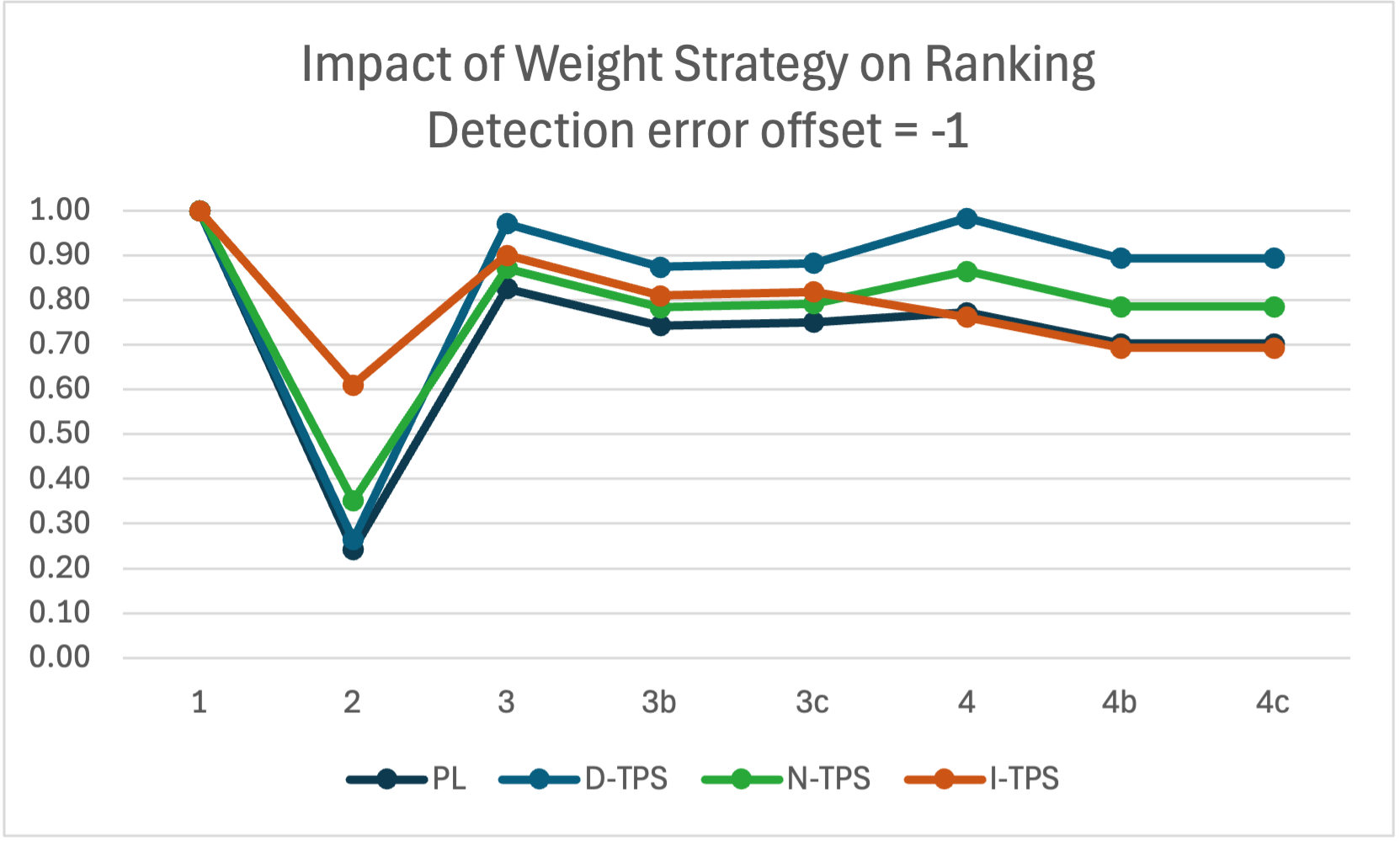}
    \caption{Impact of weight strategy on ranking, detection error offset = -1.}
    \label{fig:impactneg}
\end{figure}

First consider the flat measures. These are depicted together along with the hierarchical measure that is not weighted (PL) in Fig.~\ref{fig:flatnonweighted}. As all micro-averaged flat measures give the same scores, they are represented by $F1_{\mu}$. Model 1 has a perfect score for all; this is expected as it is ``always right.'' Model 2 has a score of 0 for every measure except $P_M$. The odd behavior seen for $P_M$ is due to some classes having no positive instances and the denominator being undefined. In this case, a score of 1 is assigned as the model does not techincally make any precision errors. Note that $P_{\mu}$ does not have this behavior. Generally, $P_M$ exhibits non-intuitive model ranking. Models 3 and 4 have the same $F1_{\mu}$ score as designed and scores are effectively the same for the variants 3b, 3c, 4b, and 4c. Note that the scores for models 3 and 4 and their variants are only non-zero because they occasionally predict correctly, not because these metrics can distinguish errors close to the truth from errors far from the truth.

$PL$ gives a slightly higher score to the more cautious model 3 when compared to the aggressive model 4, but they are close. Models with detection errors receive lower scores. ${PL}$ ranks models reasonably for the situation we describe where the minor errors made by models 3 and 4 should be distinguished from the egregious errors made by the always wrong model 2.

Next, consider the weighted hierarchical measures with $PL$ shown as a reference throughout in Figures~\ref{fig:decreasing}, ~\ref{fig:nonincreasing}, and ~\ref{fig:increasing}.
For the $L$ and $LPP$ measures for all weight strategies, model 1 does not have a perfect score; this is a noted flaw with these measures. The rest of the hierarchical measures score model 1 perfectly. 

Focusing on LPPTPS and the 0 offset (Fig.~\ref{fig:impact}), the increasing weight strategy favors the cautious model 3 over the aggressive model 4. They are scored approximately equally by the non-increasing weight strategy. The decreasing weight strategy slightly favors the aggressive model over the cautious one. Additionally, scores for decreasing weights are quite high overall due to almost full credit being given to any prediction sharing an LCA below the root node $(Rew > 0.9)$. In other words, different weight strategies can be used to favor cautious predictions, making a more general prediction when not enough information is available to be specific, over aggressive ones. As a downside, the increasing weight strategy has a strong impact on scores with standardized paths (LPPPPS and LPPTPS) driving the scores for Model 2 higher than the objectively better models 3, 3b, and 3c for both offsets in the case of LPPPPS. Due to its combination in $hF1$, it also impacts scoring here. Thus, care must be taken when designing scoring trees.

The offset that gives a greater penalty for missed and ghost detections, error = -1, shifts scores upward for models that do not make detection errors (compare Fig.~\ref{fig:impactneg} to  Fig.~\ref{fig:impact} for examples of LPPTPS scores). Model 2 sees the largest shifts due to making the most classification errors. Measures that do not standardize on path length (e.g., L and LPP) are also the most impacted while those that do are more stable. Thus, the metrics are tunable to differing impacts of classification and detection errors.
\section{Conclusions and Future Work}\label{conclusions}

This work demonstrates the development of hierarchical measures using scoring trees with adjustable weight strategies adaptable to different goals. Specifically, changing the weighting strategy can influence whether the evaluation prefers aggressive or cautious predictions. The measures are capable of handling detection errors as well with the ability to influence the evaluation of detection and classification errors. Comparisons between models remain the same when these errors are introduced at similar rates.

For most weight schemes, the hierarchical F-measure proposed worked well. However, a version of hierarchical precision, $LPPTPS$ may behave oddly with the increasing weight strategy. There is confidence, however, that  hierarchical recall, $LPP{TPS}$, remains consistent under the conditions proposed. Future exploration should be performed for macro-averaging $hP$ and $hR$ scores and comparison against other hierarchical precision and recall. Particular focus should be on what truth/prediction pairs should be included in individual class scores. So far this selection has been a class-by-class binary, but this could be better performed by rolling the descendant scores into each class in some way.   

The experimentation was limited to abstract models. Experiments on ML models with real data need to be performed to validate the methods for practical use. These should include multiple domains and more complex hierarchy structures. 

The metrics are designed for tree hierarchies. So far we've only shown that this metric can be used on tree hierarchies, but we should further expand this to be applicable to all DAGs. Other problems we have not addressed are with multi-label problems, such as in image detection hierarchies. Further work must also focus on methods for selecting weight strategies. Since it has been abstract, we've only focused on aggressive vs cautious models. Real hierarchies are needed to explore a semantics-based approach to weight selection.

\bibliographystyle{IEEEtran}
\bibliography{hs}

\end{document}